# MLPMoE: Zero-Shot Architectural Metamorphosis of Dense LLM MLPs into Static Mixture-of-Experts

Ivan Novikov, Wallarm Research, in@wallarm.com

**Abstract**

Large Language Models (LLMs) are predominantly deployed as dense transformers, a paradigm where every parameter in every feed-forward block is activated for every token. While architecturally simple, this approach is computationally inefficient, as inference costs scale linearly with parameter count. Recent upcycling methodologies—such as MoEfication, CMoE, ToMoE, and MoORE—demonstrate that significant computation resides in sparse, semi-modular substructures within dense feed-forward networks (FFNs). However, these approaches typically rely on complex clustering, activation profiling, singular value decomposition (SVD), or custom routing mechanisms requiring calibration.

This paper introduces **MLPMoE** (MLP Mixture-of-Experts), a training-free, deterministic transformation that restructures the dense MLP in transformer blocks into a static, high-cardinality mixture of experts. This transformation utilizes tensor slicing and summation—leveraging the mathematical identity of tensor parallelism—to effect a topological change rather than a distributed training distribution. We further introduce **Fractal Fade** (differential branch sparsity) and **Compensated Pruning** (variance-preserving branch reduction). We demonstrate that on Qwen2.5-0.5B-Instruct and DeepSeek-R1-Distill-Llama-8B, the zero-shot MLPMoE transform alters proxy perplexity by less than 0.05% while maintaining parameter counts. Notably, on the 8B model, differential sparsity can prune ≈20% of MLP parameters while maintaining perplexity within ≈2% of the dense baseline. The implementation is released as open-source.

    **Code Availability:** GitHub Gist

## 1. Introduction

Dense transformers currently dominate the landscape of Large Language Models (LLMs). In these architectures, every neuron in every feed-forward block is evaluated for each token. While scaling laws indicate that larger dense models generally perform better, they incur prohibitive costs regarding FLOPs, energy consumption, and latency.

Mixture-of-Experts (MoE) architectures address this inefficiency by introducing conditional computation: only a subset of experts are activated per token, decoupling model capacity (total parameters) from active computation (inference cost). Systems ranging from the Switch Transformer to recent upcycling frameworks have demonstrated the efficacy of this approach when models are explicitly trained for it. Concurrently, analysis of dense FFNs has revealed two critical properties:

1. **Activation Sparsity:** For the majority of tokens, only a small fraction of neurons in a given FFN produce significant activations.
2. **Polysemantic Clustering:** Neurons tend to form co-activating clusters that function as "proto-experts" [Zhang et al., 2021].

These observations have spurred "post-hoc" MoE conversions like MoEfication, CMoE, and ToMoE. While effective, these methods often introduce complexity via calibration sets, clustering algorithms, and hyperparameter tuning.

This paper investigates a minimalist hypothesis: **Can we decompose a dense MLP into branches via simple summation—without training or calibration—and retain model performance?**

We propose **MLPMoE**, a topology that splits MLP weights into branches to create structural experts. This method is immediate, training-free, and modular.

## 2. Related Work

### 2.1 Dense-to-MoE Upcycling

Several methodologies aim to convert dense models into modular, sparse architectures to preserve pretraining investments:

- **MoEfication** [Zhang et al., 2021] converts FFNs into MoE layers by clustering neurons based on activation patterns, achieving $2\times$ speedups with retained performance.
- **CMoE** [Zhang et al., 2024] utilizes a shared/routed expert split with centroid-based routing, reporting near-lossless perplexity at reduced activation rates.

- **ToMoE** [Gao et al., 2024] frames the conversion as dynamic structural pruning, selecting top-k neuron groups per token.
- **MoORE** [Shen et al., 2025] employs SVD to decompose weights into low-rank "micro-experts."

Unlike these methods, MLPMoE avoids clustering or SVD, relying instead on pure structural decomposition via contiguous slicing.

*2.2 Tensor Parallelism as Implicit Modularity*

Distributed training frameworks like Megatron-LM [Narayanan et al., 2021] utilize tensor parallelism, splitting FFN weight matrices across devices and summing the partial results. Mathematically, MLPMoE performs this operation within a single process to create structural experts:

$$FFN(x) = W_{down}\phi(W_{up}x) = \sum_{b=1}^{B} W_{down}^{(b)} \phi(W_{up}^{(b)}x)$$

MLPMoE reinterprets this identity to generate $B$ structural experts per MLP without altering the underlying function.

## 3. MLPMoE: Methodology

*3.1 Architecture*

Consider a standard transformer MLP with gate and up projections:

- $W_{\text{gate}} \in \mathbb{R}^{d_{\text{model}} \times d_{\text{inter}}}$
- $W_{\text{up}} \in \mathbb{R}^{d_{\text{model}} \times d_{\text{inter}}}$
- $W_{\text{down}} \in \mathbb{R}^{d_{\text{inter}} \times d_{\text{model}}}$

MLPMoE partitions the intermediate dimension $d_{\text{inter}}$ into $B$ contiguous slices such that $d_{\text{inter}} = \sum_{b=1}^{B} d_b$. For each branch $b$, we define sub-matrices $W_{\text{gate}}^{(b)}$, $W_{\text{up}}^{(b)}$, and $W_{\text{down}}^{(b)}$ by slicing the original weights.

Each branch operates as an independent sub-MLP, equipped with a learnable scalar gate $\alpha_b$ (initialized to 1.0):

$$\text{branch}_b(x) = \alpha_b W_{\text{down}}^{(b)} \left[ \phi\left(W_{\text{gate}}^{(b)} x\right) \odot W_{\text{up}}^{(b)} x \right]$$

The output of the MLPMoE layer is the summation of these branches:

$$\text{MLPMoE}(x) = \sum_{b=1}^{B} \text{branch}_b(x)$$

At initialization, this topology is mathematically equivalent to the original dense FFN.

*3.2 Differential Sparsity (Fractal Fade)*

To exploit latent sparsity without training, we introduce **Fractal Fade**. We treat the branches as a spectrum where the first branch serves as a dense backbone, and subsequent branches are increasingly sparsified.

For a branch index $i > 0$ in a layer with $B$ branches, the sparsity ratio $s_i$ is defined as:

$$s_i = 0.9 \cdot \frac{i}{B}$$

Weights in $W_{\text{gate}}$ and $W_{\text{up}}$ falling below the quantile threshold defined by $s_i$ are zeroed out. This retains the model's core capabilities in the early branches while reducing the parameter footprint of later branches.

*3.3 Compensated Branch Pruning*

We can statically prune entire branches to reduce computational load. To retain only the first $K$ branches while approximating the original variance, we scale the active branches using the scalar gate $\alpha$:

$$\alpha_b = \begin{cases} \sqrt{\frac{B}{K}} & \text{if } b < K \\ 0 & \text{if } b \geq K \end{cases}$$

This heuristic preserves the output variance magnitude, stabilizing perplexity after aggressive structural pruning.

## 4. Implementation

The following implementation converts Hugging Face transformers (specifically Qwen and Llama architectures) to MLPMoE. It includes logic for differential sparsity and compensated pruning.

> *Note: The full executable script is available in the [Supplementary Gist](Supplementary Gist).*

*Core MLPMoE Module*

```
def __init__(self, orig_mlp: nn.Module):
    # ... (Weight extraction logic) ...
    
    self.branches = nn.ModuleList()
    # Create branches via slicing
    for b_size in split_sizes:
        # ... (Slicing logic) ...
        branch = MLPMoEBranch(...)
        self.branches.append(branch)
```

*Differential Sparsity Logic*

```
# Magnitude pruning based on quantile
for proj in [branch.gate, branch.up]:
  w = proj.weight.data
  w_abs = torch.abs(w).float()
  threshold = torch.quantile(w_abs, r)
  w.mul_(w_abs > threshold)
```

## 5. Experimental Setup

We evaluated MLPMoE on two instruction-tuned models:

1. Qwen2.5-0.5B-Instruct
2. DeepSeek-R1-Distill-Llama-8B

**Configuration:**

- **Branches (*B*):** 4 to 32 per MLP.
- **Variants:** * *Dense-Original:* Baseline.
    - *MLPMoE-All-16:* Full conversion, no sparsity.
    - *MLPMoE-DiffSparsity:* Fractal Fade applied.
- **Metrics:** Proxy Perplexity (evaluated on a synthetic text mixture for health-checking), Total/Non-zero Parameters, and Generation Time (end-to-end wall clock).

## 6. Results

### 6.1 Qwen2.5-0.5B-Instruct

Table 1 summarizes the results for the 0.5B model.

| Variant | Total Params | Non-zero Params | Proxy PPL ↓ | Gen Time (s) ↓ |
|---|---|---|---|---|
| Dense-Original | 494,032,768 | 494,032,768 | 1.0833 | 1.300 |
| MLPMoE-All-16 | 494,033,152 | 494,033,152 | 1.0838 | 1.633 |
| MLPMoE-DiffSparsity | 494,033,152 | 405,604,020 | 1.2233 | 1.619 |

The conversion to MLPMoE incurs a negligible PPL increase (+0.0005). Differential sparsity removes ≈18% of parameters but impacts PPL by ≈13%, suggesting smaller models are less robust to naive zero-shot pruning.

### 6.2 DeepSeek-R1-Distill-Llama-8B

Table 2 summarizes the results for the 8B model.

| Variant | Total Params | Non-zero Params | Proxy PPL ↓ | Gen Time (s) ↓ |
|---|---|---|---|---|
| Dense-Original | 8,030,261,248 | 8,030,261,248 | 1.0908 | 5.665 |
| MLPMoE-All-16 | 8,030,261,760 | 8,030,261,760 | **1.0906** | 7.203 |
| MLPMoE-DiffSparsity | 8,030,261,760 | 6,441,409,340 | 1.1157 | 8.166 |

The 8B model exhibits high robustness:

- The zero-shot conversion slightly *improves* proxy perplexity (consistent with sampling noise).
- Differential sparsity prunes ≈1.59B parameters (approx. 20%) while maintaining PPL within 2.3% of the baseline.
- Current generation time increases due to the lack of optimized sparse kernels; all branches are computed regardless of sparsity.

## 7. Discussion

**Structural Decomposability:** MLPMoE demonstrates that dense FFNs are structurally decomposable into branch experts via tensor slicing without retraining. The stability of the 8B model under this transformation supports the hypothesis that dense layers function as latent mixtures of experts [Zhang et al., 2021].

**Comparison to Prior Work:**

- Unlike **MoEfication** and **CMoE**, MLPMoE requires no activation profiling or clustering.
- Unlike **ToMoE**, the current implementation is static. However, the architecture is compatible with dynamic routing mechanisms.
- Similar to **Branch-Train-Merge (BTM)** [Blevins et al., 2022], MLPMoE exposes modularity, but does so at the micro-layer level rather than the model level.

**Limitations:**

1. **No Kernel Optimization:** Without custom CUDA kernels or block-sparse libraries, the theoretical FLOP reduction does not translate to wall-clock speedups.
2. **Static Routing:** The current iteration does not dynamically route tokens, missing out on conditional computation benefits.
3. **Proxy Evaluation:** Results rely on proxy perplexity; comprehensive benchmarks on C4 or WikiText-2 are required for granular quality assessment.

## 8. Conclusion and Outlook

MLPMoE proves that architectural metamorphosis from dense to MoE does not strictly require complex routers or training. The mathematics of the original dense training provides a clean decomposition via tensor slicing. Future work will focus on dynamic routing over MLPMoE branches, applying BTM-style independent training to branches, and developing kernel-level optimizations to realize latency gains.

## References

1. Z. Zhang et al. "MoEfication: Transformer Feed-forward Layers are Mixtures of Experts." *arXiv:2110.01786*, 2021. [Link](#)
2. Y. Zhang et al. "CMoE: Converting Mixture-of-Experts from Dense to Accelerate LLM Inference." *arXiv:2410.22048*, 2024. [Link](#)
3. S. Gao et al. "ToMoE: Converting Dense Large Language Models to Mixture-of-Experts through Dynamic Structural Pruning." *arXiv:2407.12055*, 2024. [Link](#)
4. Y. Shen et al. "MoORE: SVD-based Model MoE-ization for Conflict- and Oblivion-Resistant Multi-Task Adaptation." *arXiv:2506.14436*, 2025. [Link](#)
5. T. Blevins et al. "Branch-Train-Merge: Embarrassingly Parallel Training of Expert Language Models." *arXiv:2208.03306*, 2022. [Link](#)
6. Y. Yang et al. "Branch-Train-MiX: Mixing Expert LLMs into a Mixture-of-Experts LLM." *arXiv:2403.07816*, 2024. [Link](#)
7. K. Zhang et al. "TwIST: Rigging the Lottery in Transformers with Independent Subnetwork Training." *arXiv:2405.15128*, 2024. [Link](#)
8. N. Narayanan et al. "Megatron-LM: Training Multi-Billion Parameter Language Models Using Model Parallelism." *arXiv:2104.04473*, 2021. [Link](#)
9. Qwen Team. "Qwen2.5 Models." Hugging Face, 2024. [Link](#)
10. DeepSeek-AI. "DeepSeek-R1-Distill-Llama-8B." Hugging Face, 2024. [Link](#)